\documentclass{article}
\usepackage{spconf,amsmath,graphicx}
\usepackage{multirow}
\usepackage{cite}

\title{Learning Multiple Explainable and Generalizable Cues for Face Anti-spoofing}
\name{Ying Bian$^\star$, Peng Zhang$^\star$\thanks{Equal contribution.}, Jingjing Wang, Chunmao Wang, Shiliang Pu\sthanks{Corresponding author: Shiliang Pu (pushiliang.hri@hikvision.com).}}
\address{Hikvision Research Institute, China}
\begin{document}
%
\maketitle
\begin{abstract}
Although previous CNN based face anti-spoofing methods have achieved promising performance under intra-dataset testing, they suffer from poor generalization under cross-dataset testing. The main reason is that they learn the network with only binary supervision, which may learn arbitrary cues overfitting on the training dataset. To make the learned feature explainable and more generalizable, some researchers introduce facial depth and reflection map as the auxiliary supervision. However, many other generalizable cues are unexplored for face anti-spoofing, which limits their performance under cross-dataset testing. To this end, we propose a novel framework to learn multiple explainable and generalizable cues (MEGC) for face anti-spoofing. Specifically, inspired by the process of human decision, four mainly used cues by humans are introduced as auxiliary supervision including the boundary of spoof medium, moir\'e pattern, reflection artifacts and facial depth in addition to the binary supervision. To avoid extra labelling cost, corresponding synthetic methods are proposed to generate these auxiliary supervision maps. Extensive experiments on public datasets validate the effectiveness of these cues, and state-of-the-art performances are achieved by our proposed method.
\end{abstract}
\section{Introduction}
\label{sec:intro}
Currently, face anti-spoofing has become a crucial part to guarantee the security of face recognition systems and drawn increasing attention in the face recognition community. Previous methods mainly extract handcrafted features such as color \cite{2017Face}, texture and distortion cues \cite{2015Face} for face anti-spoofing. However these methods are vulnerable to illumination variations and scene changes.

As deep learning has proven to be effective in many computer vision problems, many researchers turn to employ CNNs to extract more discriminative features \cite{2014Learn, 2018Face, 2018De-Spoofing}, and show significant improvement over the conventional ones. These methods treat face anti-spoofing as a binary classification problem and train the network with only softmax loss. A CNN with binary supervision might discover arbitrary cues to separate the two classes without explanation, which causes overfitting on the training dataset. When the learned cues change or even disappear during testing, these models would fail to distinguish spoof vs. live faces and achieve poor generalization performance under cross-dataset testing. Therefore, it is desirable to learn explainable and generalizable cues for face anti-spoofing.

\begin{figure}[t!]
	\centering
	\includegraphics[width=1\linewidth]{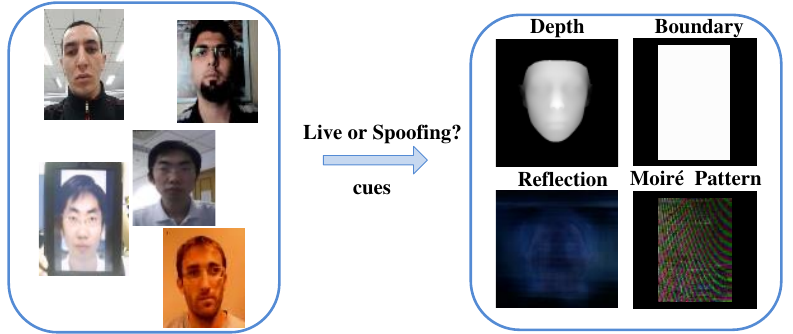}
	\caption{Face anti-spoofing can be regarded as a binary classification (live or spoofing) problem, which relies on the intrinsic cues such as depth, reflection, boundary and moir\'e pattern.}
	\label{fig:intro}
\end{figure}

To achieve this goal, Liu et al. \cite{2018Learning} regard live faces have face-like depth, while faces in print or replay attacks have flat or planar depth. Therefore they utilize depth as auxiliary information to supervise both live and spoof faces. Considering light rays that are reflected from a surface of spoof medium may cause the reflection artifacts in recaptured images, the reflection map \cite{2019basn} is used as additional auxiliary supervision for more robust feature learning. However, only limited cues are leveraged in these approaches, and many other generalizable cues (such as moir\'e pattern, boundary of spoof medium, etc.) are discarded for face anti-spoofing, which limits their performance under cross-dataset testing. Therefore, more generalizable cues are desirable to be explored to improve the robustness under severe variations.

When a human is distinguishing spoof vs. live faces, the following four main artifacts are usually leveraged. Firstly, the boundary of the spoof medium, such as the screen border of the phone and computer, or the boundary of the printed photographs is easily spotted. Secondly, there exists obvious moir\'e pattern under replay-attack due to the aliasing caused by different frequencies of capture devices. Thirdly, reflection artifacts may be caused by the reflection from a surface of spoof medium. Finally, facial depth difference between live and spoofing faces is also a cue as most spoofing faces are broadcasted in plane presentation attack instruments.

Inspired by the way humans distinguish spoof vs. live faces, we propose a novel framework to learn multiple explainable and generalizable cues for face anti-spoofing. Specifically, the network is trained end-to-end with boundary of spoof medium, moir\'e pattern, reflection artifacts and facial depth as auxiliary supervision in addition to the binary supervision. These extracted cues are visualized in Fig.\ref{fig:intro}. Due to the expensive cost for labelling these cues, we propose synthetic methods to generate the corresponding maps.
\section{PROPOSED METHOD}
\label{sec:majhead}

\begin{figure*}[ht!]
	\centering
	\includegraphics[width=1\linewidth]{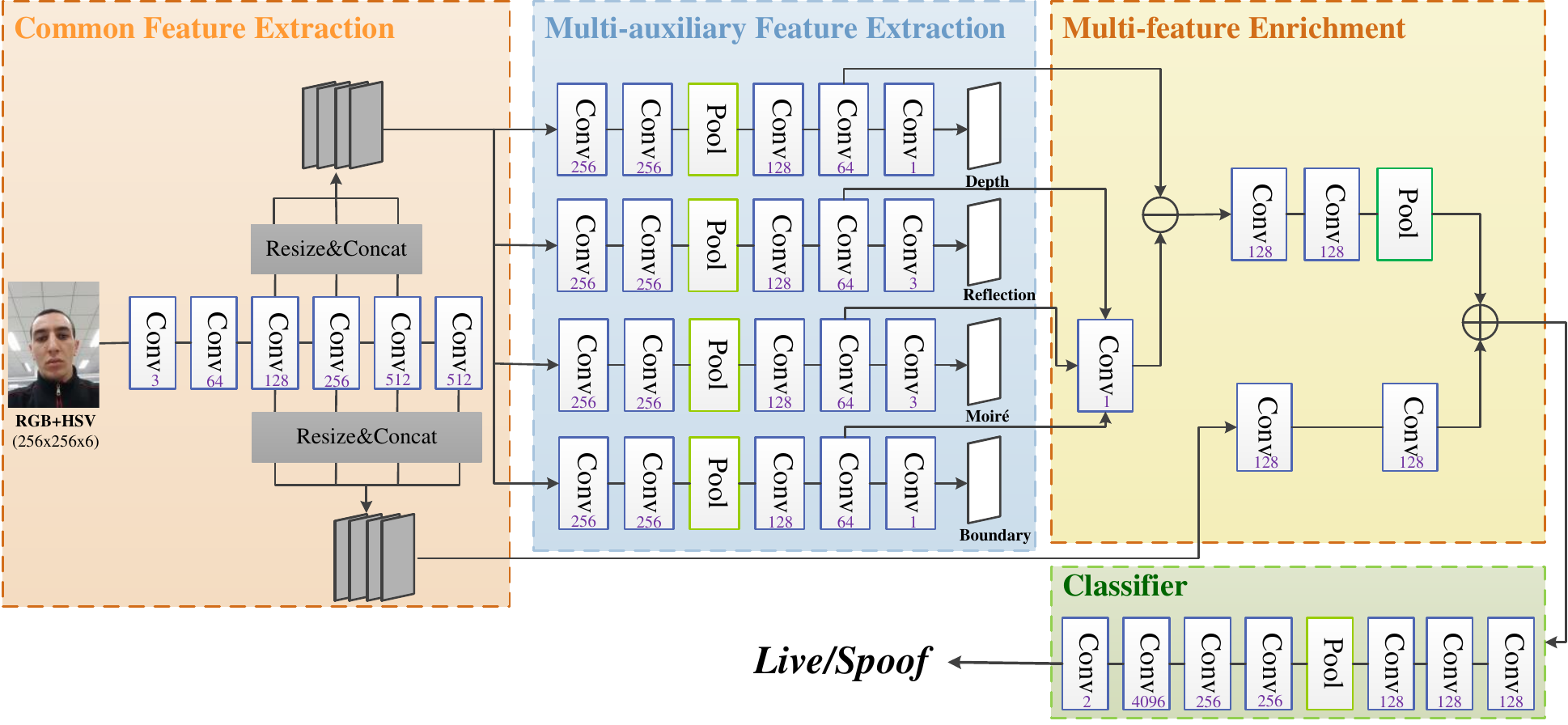}
	\caption{The architecture of our MEGC framework. Individual numbers indicate the channel numbers of feature maps.}
	\label{fig:overview}
\end{figure*}

In order to learn the proposed explainable and generalizable cues for face anti-spoofing, we need to get the auxiliary supervision maps including the ones of the boundary of spoof medium, moir\'e pattern, reflection artifacts and facial depth. The reflection and depth maps are extracted as \cite{2019basn}, while the boundary and moir\'e maps are generated using our proposed synthetic methods. In the section, we first introduce the methods to generate the boundary and moir\'e maps, and then elaborate the proposed MEGC framework as illustrated in Fig.\ref{fig:overview}.

\subsection{Extracting Moir\'e Map}
\label{ssec:Moire}
When a fine pattern on the subject meshes with the pattern on the imaging chip of the shooting camera, the moir\'e pattern occurs. It is inevitable to get moir\'e pattern in the relay-attack, since a screen is the subject photographed. Therefore, moir\'e pattern is a strong generalizable cue under relay-attack.

\begin{figure}[h]
	\centering
	\includegraphics[width=1\linewidth]{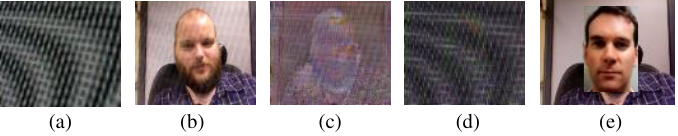}
	\caption{(a) generated moir\'e pattern, (b) generated moir\'e image, (c) the residual moir\'e map learned by demoir\'eing method MRGAN\cite{2021mrgan}, (d) moir\'e map estimated by our network, (e) synthetic image with boundary.}
	\label{fig:moire_boundary}
\end{figure}

Directly labelling the moir\'e pattern is intractable, we need algorithms to extract it automatically. At present, there is no model to directly estimate the moir\'e map of an input image. As the process of generating moir\'e pattern is known, we can use different interference fringes with similar frequency to generate moir\'e pattern physically, and add it into an image without moir\'e pattern to get a corresponding pair of input image and its moir\'e map. Another way is to leverage the existing mature demoir\'eing methods to get an output image without moir\'e pattern given the input image with moir\'e pattern. Then subtracting the output image from the input image, we can obtain the corresponding moir\'e map. However the first method only uses the live images without moir\'e pattern and discards the various spoofing images with moir\'e pattern, which limits its performance. While the second method focuses on removing the moir\'e pattern to make the output image have no moir\'e pattern visually, which often leads to the residual image contains some image content due to over-removing as shown in Fig.\ref{fig:moire_boundary} (c). This noising moir\'e map hinders the feature learning and leads to performance degradation.

\begin{figure}[h]
	\centering
	\includegraphics[width=1\linewidth]{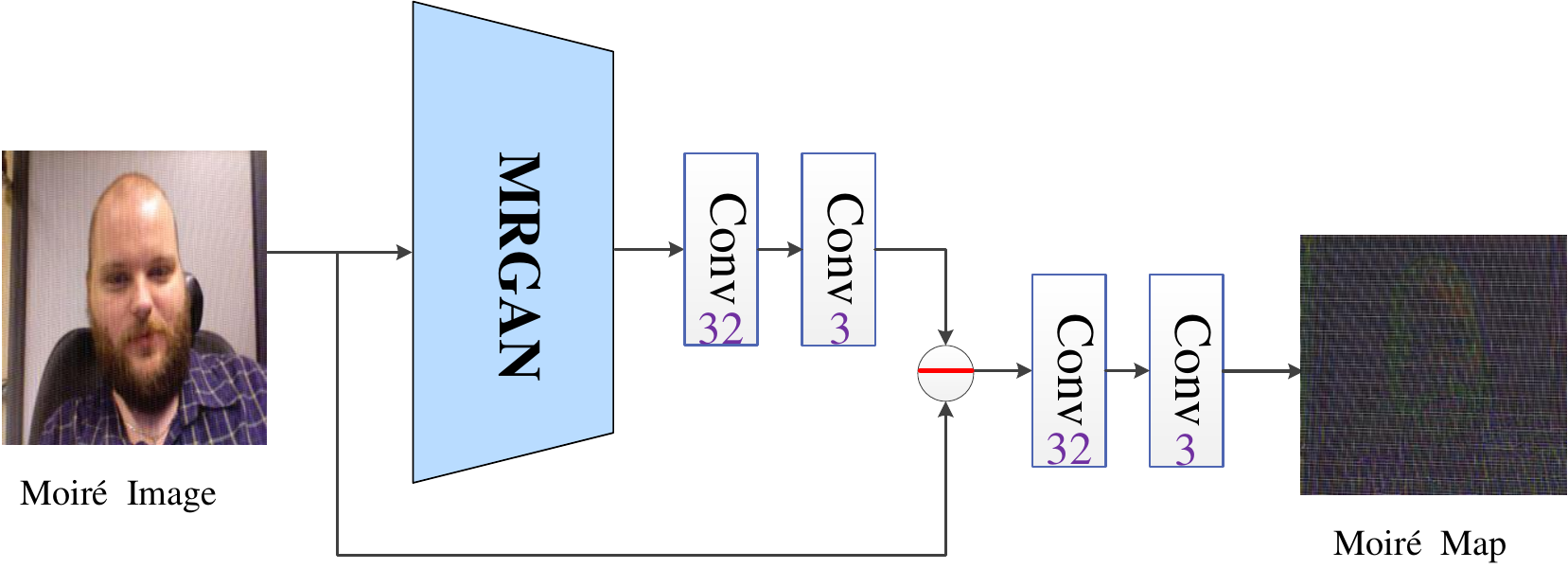}
	\caption{The network architecture for moir\'e map extraction.}
	\label{fig:moire}
\end{figure}

In order to solve this problem, we propose a network to estimate the moir\'e map of an input image as shown in Fig.\ref{fig:moire}. In the training phase, we use the above mentioned first method to get corresponding pairs of input images and their moir\'e maps. The images are as input for the network, and the corresponding moir\'e maps are as supervision. To reduce the difficulty of learning, the final moir\'e map learning is based on the above mentioned second method. That is, we first use the image demoir\'eing method to get the residual image and then refine it to get the final moir\'e map. The image demoir\'eing part is based on the SOTA demoir\'eing method MRGAN\cite{2021mrgan}. We use a trained MRGAN model\cite{2021mrgan} to initialize the parameters of the upper half branch of our network and its parameters are fixed during training, while it is followed by two learnable 3*3 convolution layers for adaptation. The image demoir\'eing part is also followed by two 3$\times$3 convolution layers to refine the moir\'e map. During testing, we use the trained network to extract the moir\'e maps of replay-attack images which are used as moir\'e cues to train the network in our MEGC framework. Moir\'e maps are set to all zeros for live faces, and we don't perform gradient back prorogation on spoofing samples which don't belong to replay spoofing types when training the auxiliary moir\'e part in our MEGC framework.

\subsection{Generating Boundary Map }
\label{ssec:Boundary}
During recapturing, the boundary of the spoof medium, such as the screen border of the phone and computer, or the boundary of the printed photograph is often captured by the camera due to the larger field of view. Therefore, the boundary of the spoof medium is another strong generalizable cue for face anti-spoofing. 

To label the boundary is labor-consuming, we propose a synthetic method to generate pairs of spoofing images with boundaries and their corresponding boundary maps.
Firstly, we randomly select a live sample and a spoofing sample. Then the face in the spoofing sample is cut out and pasted to the live sample to replace the live face. In this way, we can get spoofing samples with known boundaries as shown in Fig.\ref{fig:moire_boundary} (e). To generate the corresponding boundary map, the values inside the boundary are set to one, the ones outside the boundary are set to zero as shown in Fig.\ref{fig:intro}. As for live faces, boundary maps are set to all zeros. This boundary map will act as an auxiliary supervision to help the network in our MEGC framework to learn the boundary cues. For the original spoofing images, as we don't know their boundary information, these samples are not used to train the boundary part of the network.

\subsection{MEGC Framework}
\label{ssec:subhead}
The proposed MEGC framework consists of four main modules, i.e., common feature extraction (backbone network), multi-auxiliary feature extraction (MAFE), multi-feature enrichment (MFE), and classifier as shown in Fig.\ref{fig:overview}. To make a fair comparison with BASN\cite{2019basn}, we use the same backbone network and classifier as BASN. Feature maps of conv3, conv4, conv5 from the backbone network are resized to the fixed size of 64$\times$64, and are then concatenated to be passed to the MAFE. Feature maps of conv3, conv4, conv5, and conv6 of are resized to the size of 16$\times$16 and are concatenated to be passed to the MFE.

\textbf{Multi-auxiliary Feature Extractor}. MAFE consists of four auxiliary feature extractors, including depth feature extractor, reflection feature extractor, moir\'e feature extractor and boundary feature extractor.
The depth and reflection feature extractors are the same as the ones in BASN\cite{2019basn}. As for the moir\'e and boundary feature extractors, we get the ground truth maps using the methods proposed in section \ref{ssec:Moire} and \ref{ssec:Boundary} respectively with size of 32$\times$32. 
Given an input face image I, the MAFE predicts the depth map $D_{pre}$, reflection map $R_{pre}$, moir\'e map $M_{pre}$ and boundary map $B_{pre}$.
The loss functions can be formulated as:
\begin{align}
	&\mathcal{L}_{D} = \frac{1}{N}\sum_{i \in N}^{}||D_{pre}(i)-D_{gt}(i)||^2_{2} \\
	&\mathcal{L}_{R} = \frac{1}{N}\sum_{i \in N}^{}||R_{pre}(i)-R_{gt}(i)||^2_{2} \\
    &\mathcal{L}_{M} = \frac{1}{N}\sum_{i \in N}^{}||M_{pre}(i)-M_{gt}(i)||^2_{2} \\
	&\mathcal{L}_{B} = \frac{1}{N}\sum_{i \in N}^{}||B_{pre}(i)-B_{gt}(i)||^2_{2}
\end{align}
where, $D_{gt}$, $R_{gt}$, $M_{gt}$ and $B_{gt}$ denote ground truth depth map, reflection map,
moir\'e map and boundary map respectively. $N$ is the batch size. 
Finally, the overall loss function is $\mathcal{L}_{overall} =\mu * \mathcal{L}_{cls} + \lambda * (\mathcal{L}_{D} + \mathcal{L}_{R} + \mathcal{L}_{B}+ \mathcal{L}_{M})$, where $\mu$ and $\lambda$ denote the weight of each loss functions.

\textbf{Multi-feature Enrichment}. MFE enriches the feature representations by fusing feature maps from MAFE and the backbone network. Finally, the fused feature map will go through the binary classifier. Different from BASN\cite{2019basn}, we first fuse the reflection, moir\'e and boundary features from MAFE as the spoofing feature map. Then the spoofing feature map is subtracted from the depth feature map.

\section{EXPERIMENTS}
\label{sec:format}
\subsection{Experimental Setup}
\textbf{Datasets}. Two public face anti-spoofing datasets are utilized to evaluate the effectiveness of our method: Replay-Attack\cite{2012Replay} (denoted as R) and CASIA-MFSD\cite{Zhang2012A} (denoted as C). We select one dataset as source domain for training and the remaining one as target domain for cross-testing. Thus, we have two cross-testing tasks in total. Following \cite{2019Multi}, the Half Total Error Rate (HTER) is used as the evaluation metric.

\textbf{Implementation Details}. The size of face image is 256$\times$256$\times$6 with both the RGB and HSV channels. The face boxes are detected with open source toolbox Dlib. To expose the boundary, the face boxes are expanded by 1 times in size. Other hyperparameters $\mu$, $\lambda$ are set to 10, 0.1 respectively. For every training epoch, the ratio of positive and negative images is 1:1.

\subsection{Ablation Study}
\label{ssec:subhead}
To verify the effectiveness of the learned spoofing cues, we discard one of the spoofing cues in turn to get our methods without reflection, moir\'e and boundary cues, which are denoted as Ours\_wo/R, Ours\_wo/M and Ours\_wo/B respectively. The results of Tab.\ref{tab:ablation} show that the performances of these methods decrease in different degrees which validates the effectiveness of each one of the proposed spoofing cues. It is also worth noting that different cues have different impact on different datasets. The performance of Ours\_wo/M is the worst on the Replay-Attack which shows the moir\'e cue is the most robust cue for Replay-Attack. Similarly, we can get the boundary cue is the most robust cue for CASIA-MFSD. Therefore, learning multiple generalizable cues can improve the robustness of the model under cross-testing.

To verify the effectiveness of our moir\'e extracting method, we compare our method with the two mentioned extracting methods in subsection \ref{ssec:Moire}. The first one adds synthetic moir\'e (as shown in Fig.\ref{fig:moire_boundary} (a)) into the live images to get the training images with moir\'e, which is denoted as Ours\_w/moir\'e1. The second one leverages the demoir\'eing method to get the residual moir\'e map , which is denoted as Ours\_w/moir\'e2. The results of Tab.\ref{tab:ablation} show the effectiveness of our proposed moir\'e extracting method. Ours\_w/moir\'e1 discards the real spoofing images with moir\'e pattern, while the residual moir\'e map learned by Ours\_w/moir\'e2 contains some image content as shown in Fig.\ref{fig:moire_boundary} (c), which hinders their performances. Our method can learn clean moir\'e map as shown in Fig.\ref{fig:moire_boundary} (d).
\begin{table}[t!]
	\setlength{\abovecaptionskip}{0cm} 
	\setlength{\belowcaptionskip}{-0.2cm}
	\small
	\caption{Ablation study on cross-dataset testing.}
	\centering
	\begin{center}
		\begin{tabular}{|c|c|c|c|c|}
			\hline
			\multirow{2}{*}{\textbf{Methods}} &Train &Test &Train &Test	\\ \cline{2-5}
			&\begin{tabular}[c]{@{}l@{}}C\end{tabular} &\begin{tabular}[c]{@{}l@{}}R\end{tabular} &\begin{tabular}[c]{@{}l@{}}R\end{tabular} &\begin{tabular}[c]{@{}l@{}}C\end{tabular} \\ \hline 		
			Ours &\multicolumn{2}{c|}{\textbf{20.2}} &\multicolumn{2}{c|}{\textbf{27.9}} \\ \hline	
			Ours\_wo/R &\multicolumn{2}{c|}{$25.7$} &\multicolumn{2}{c|}{$35.2$} \\ \hline
			Ours\_wo/M &\multicolumn{2}{c|}{$29.0$} &\multicolumn{2}{c|}{$34.1$} \\ \hline
			Ours\_wo/B &\multicolumn{2}{c|}{$23.7$} &\multicolumn{2}{c|}{$37.2$} \\ \hline
			Ours\_w/moir\'e1 &\multicolumn{2}{c|}{$27.9$} &\multicolumn{2}{c|}{$39.6$} \\ \hline
			Ours\_w/moir\'e2 &\multicolumn{2}{c|}{$30.8$} &\multicolumn{2}{c|}{$39.8$} \\ \hline
		\end{tabular}
	\end{center}
	\label{tab:ablation}
\end{table}

\subsection{Comparison with State-of-the-Art Methods}
\label{ssec:subhead}
In this subsection, we compare the proposed MEGC with previous state-of-the-art methods. The competitive approaches include LBP-TOP\cite{2014dynamic}, Spectral cubes\cite{Spectralcubes}, LBP\cite{LBP}, Color Texture\cite{2017Face}, CNN\cite{2014Learn}, STASN\cite{STASN}, FaceDe-S\cite{fDes}, Auxiliary\cite{2018Learning}, BASN\cite{2019basn}, BCN\cite{bcn}. As shown in Tab.\ref{tab:SOTA}, the bold type indicates the best performance, and the under-line type indicates the second best performance. Our method outperforms the baseline method BASN\cite{2019basn}, which verifies the effectiveness of the two extra introduced cues: moir\'e and boundary. Our approach achieves the best overall performance, which verifies the effectiveness of proposed multiple generalizable cues learning. It is notable that our method is slightly worse than BCN\cite{bcn} on the Replay-Attack. However, BCN uses more sophisticated network and introduces other cues, such as surface texture cues. These cues are compatible with our method, and combination of them can further improve the performance. We leave it in the future work.

\begin{table}[t!]
	\setlength{\abovecaptionskip}{0cm} 
	\setlength{\belowcaptionskip}{-0.2cm}
	\small
	\caption{Comparison to SOTA methods.}
	\small
	\centering
	\begin{center}
		\begin{tabular}{|c|c|c|c|c|}
		\hline
		\multirow{2}{*}{\textbf{Methods}} &Train &Test &Train &Test	\\ \cline{2-5}
			&\begin{tabular}[c]{@{}l@{}}C\end{tabular} &\begin{tabular}[c]{@{}l@{}}R\end{tabular} &\begin{tabular}[c]{@{}l@{}}R\end{tabular} &\begin{tabular}[c]{@{}l@{}}C\end{tabular} \\ \hline 			
		LBP\-TOP\cite{2014dynamic} &\multicolumn{2}{c|}{$49.7$} &\multicolumn{2}{c|}{$60.6$} \\ \hline
		Spectral cubes\cite{Spectralcubes} &\multicolumn{2}{c|}{$34.4$} &\multicolumn{2}{c|}{$50.0$} \\ \hline
		LBP\cite{LBP} &\multicolumn{2}{c|}{$47.0$} &\multicolumn{2}{c|}{$39.6$} \\ \hline
		Color Texture\cite{2017Face} &\multicolumn{2}{c|}{$30.3$} &\multicolumn{2}{c|}{$37.7$} \\ \hline
		CNN\cite{2014Learn} &\multicolumn{2}{c|}{$48.5$} &\multicolumn{2}{c|}{$45.5$} \\ \hline
		STASN\cite{STASN} &\multicolumn{2}{c|}{$31.5$} &\multicolumn{2}{c|}{$30.9$} \\ \hline
		FaceDe-S\cite{fDes} &\multicolumn{2}{c|}{$28.5$} &\multicolumn{2}{c|}{$41.1$} \\ \hline
		Auxiliary\cite{2018Learning} &\multicolumn{2}{c|}{$27.6$} &\multicolumn{2}{c|}{\underline{28.4}} \\ \hline
		BASN\cite{2019basn} &\multicolumn{2}{c|}{$23.6$} &\multicolumn{2}{c|}{$29.9$} \\ \hline
		BCN\cite{bcn} &\multicolumn{2}{c|}{\textbf{16.6}} &\multicolumn{2}{c|}{$36.4$} \\ \hline
		\textbf{Ours} &\multicolumn{2}{c|}{\underline{20.2}} &\multicolumn{2}{c|}{\textbf{27.9}} \\ \hline
		\end{tabular}
	\end{center}
	\label{tab:SOTA}
\end{table}

\section{CONCLUSION}
\label{sec:format}
In this paper, we propose a novel framework to learn multiple explainable and generalizable cues for face anti-spoofing. Moir\'e pattern and boundary of spoof medium are introduced to improve the generalization capacity. Two synthetic methods are proposed to generate the corresponding maps to avoid the expensive cost for labelling. Extensive experiments show the effectiveness of these cues, and state-of-the-art performances are achieved.



\bibliographystyle{IEEEbib}
\bibliography{refs}

\begin{thebibliography}{10}

\bibitem{2017Face}
Zinelabidine Boulkenafet, Jukka Komulainen, and Abdenour Hadid,
\newblock ``Face spoofing detection using colour texture analysis,''
\newblock {\em IEEE Transactions on Information Forensics and Security}, vol.
  11, no. 8, 2016.

\bibitem{2015Face}
Wen, D., Han, H., Jain, and A.K.,
\newblock ``Face spoof detection with image distortion analysis,''
\newblock {\em IEEE Transactions on Information Forensics and Security}, vol.
  10, no. 4, pp. 746--761, 2015.

\bibitem{2014Learn}
Jianwei Yang, Zhen Lei, and Stan~Z Li,
\newblock ``Learn convolutional neural network for face anti-spoofing,''
\newblock in {\em arXiv preprint arXiv:1408.5601}, 2014.

\bibitem{2018Face}
Yousef Atoum, Yaojie Liu, Amin Jourabloo, and Xiaoming Liu,
\newblock ``Face anti-spoofing using patch and depth-based cnns,''
\newblock in {\em Proceedings of the IEEE International Joint Conference on
  Biometrics (IJCB)}, 2017.

\bibitem{2018De-Spoofing}
Jourabloo, A., Liu, and X.,
\newblock ``Face de-spoofing: Anti-spoofing via noise modeling,''
\newblock in {\em Proceedings of the European Conference on Computer Vision
  (ECCV)}, 2018.

\bibitem{2018Learning}
Yaojie Liu, Amin Jourabloo, and Xiaoming Liu,
\newblock ``Learning deep models for face anti-spoofing: Binary or auxiliary
  supervision,''
\newblock in {\em Proceedings of the IEEE Conference on Computer Vision and
  Pattern Recognition (CVPR)}, 2018.

\bibitem{2019basn}
Kim, T., Kim, Y., Kim, I., Kim, and D.,
\newblock ``Basn: Enriching feature representation using bipartite auxiliary
  supervisions for face anti-spoofing,''
\newblock {\em International Conference on Computer Vision Workshops (ICCVW)},
  2019.

\bibitem{2021mrgan}
Huanjing Yue, Yijia Cheng, Fanglong Liu, and Jingyu Yang,
\newblock ``Unsupervised moiré pattern removal for recaptured screen images,''
\newblock {\em Neurocomputing}, vol. 456, pp. 352--363, 2021.

\bibitem{2012Replay}
Ivana Chingovska, Andr\'e Anjos, and Sebastien Marcel,
\newblock ``On the effectiveness of local binary patterns in face
  antispoofing,''
\newblock in {\em Proceedings of the International Conference of Biometrics
  Special Interest Group (BIOSIG)}, 2012.

\bibitem{Zhang2012A}
Zhiwei Zhang, Junjie Yan, Sifei Liu, Zhen Lei, Dong Yi, and Stan~Z Li,
\newblock ``A face antispoofing database with diverse attacks,''
\newblock in {\em Proceedings of the IEEE International Conference on
  Biometrics (ICB)}, 2012.

\bibitem{2019Multi}
Rui Shao, Xiangyuan Lan, Jiawei Li, and Pong~C. Yuen,
\newblock ``Multi-adversarial discriminative deep domain generalization for
  face presentation attack detection,''
\newblock in {\em Proceedings of the IEEE Conference on Computer Vision and
  Pattern Recognition (CVPR)}, 2019.

\bibitem{2014dynamic}
Freitas Pereira, T., Komulainen, J., Anjos, A., De~Martino, J., Hadid, A.,
  Pietikäinen, M., Marcel, and S.,
\newblock ``Face liveness detection using dynamic texture,''
\newblock {\em Eurasip Journal on Image and Video Processing}, vol. 1, no. 2,
  2014.

\bibitem{Spectralcubes}
Allan Pinto, Helio Pedrini, William~Robson Schwartz, and Anderson Rocha,
\newblock ``Face spoofing detection through visual codebooks of spectral
  temporal cubes,''
\newblock {\em IEEE Transactions on Image Processing}, vol. 24, pp. 12, 2015.

\bibitem{LBP}
Boulkenafet, Z., Komulainen, J., Hadid, and A.,
\newblock ``Face anti-spoofing based on color texture analysis,''
\newblock in {\em International Conference on Image Processing (ICIP)}, 2015,
  pp. 2636--2640.

\bibitem{STASN}
Xiao Yang, Wenhan Luo, Linchao Bao, Yuan Gao, Dihong Gong, Shibao Zheng,
  Zhifeng Li, and Wei Liu,
\newblock ``Face antispoofing: Model matters, so does data,''
\newblock in {\em Proceedings of the IEEE Conference on Computer Vision and
  Pattern Recognition (CVPR)}, 2019.

\bibitem{fDes}
Amin Jourabloo, Yaojie Liu, and Xiaoming Liu,
\newblock ``Face despoofing: Anti-spoofing via noise modeling,''
\newblock in {\em Proceedings of the European Conference on Computer Vision
  (ECCV)}, 2018.

\bibitem{bcn}
Zitong Yu, Xiaobai Li, Xuesong Niu, Jingang Shi, and Guoying Zhao,
\newblock ``Face anti-spoofing with human material perception,''
\newblock in {\em Proceedings of the European Conference on Computer Vision
  (ECCV)}, 2020.

\end{thebibliography}

\end{document}